# Capacity, Fidelity, and Noise Tolerance of Associative Spatial-Temporal Memories Based on Memristive Neuromorphic Networks

Dmitri Gavrilov*, *Member, IEEE*, Dmitri Strukov**, *Senior Member, IEEE*,
and Konstantin K. Likharev*, *Fellow, IEEE*

***Abstract*—We have calculated the key characteristics of associative (content-addressable) spatial-temporal memories based on neuromorphic networks with restricted connectivity – "CrossNets". Such networks may be naturally implemented in nanoelectronic hardware using hybrid CMOS/memristor circuits, which may feature extremely high energy efficiency, approaching that of biological cortical circuits, at much higher operation speed. Our numerical simulations, in some cases confirmed by analytical calculations, have shown that the characteristics depend substantially on the method of information recording into the memory. Of the four methods we have explored, two look especially promising – one based on the quadratic programming, and the other one being a specific discrete version of the gradient descent. The latter method provides a slightly lower memory capacity (at the same fidelity) then the former one, but it allows local recording, which may be more readily implemented in nanoelectronic hardware. Most importantly, at the synchronous retrieval, both methods provide a capacity higher than that of the well-known Ternary Content-Addressable Memories with the same number of nonvolatile memory cells (e.g., memristors), though the input noise immunity of the CrossNet memories is somewhat lower.**

***Index Terms*—Spatial-temporal memories, associative memories, nanoelectronics, neuromorphic networks, hybrid circuits, memristors, CrossNets, capacity, noise tolerance**

## I. Introduction

ASSOCIATIVE spatial-temporal memories (ASTM), which record a time sequence of similarly-formatted spatial patterns, and then may reproduce the whole sequence upon the input of just one of these patterns (possibly, contaminated by noise), are valuable parts of cognitive systems. Indeed, we all know how a few overheard notes trigger our memory of an almost-forgotten tune. (Such observations have been confirmed by neurobiological studies – see, e.g., the review [1].) Another example (which also gives a very natural language for the description of such sequences, used in this paper), is a reproduction of a movie, triggered by the input of just its one, possibly incomplete or partly corrupted, frame.

The recent fast progress of mixed-signal nanoelectronic hardware, in particular of hybrid CMOS/memristor circuits (see, e.g., the reviews [2, 3]), may enable ASTMs with extremely high speed and energy efficiency. One option here is to use the so-called Ternary Content-Addressable Memory (T-CAM) architecture - see, e.g., Ref. [4]. Indeed, as was discussed in Ref. [5], the memristor version of such a memory requires just 2 memristors per cell. As a result, the total number $n$ of these devices in an associative memory holding $Q$ spatial patterns ("frames"), of $N$ bits ("binary pixels") each, is just $2NQ$, i.e. is only twice larger than that necessary for the usual resistive memory (with no noise correction ability).

In this paper, we will show that these hardware costs may be reduced significantly using the hybrid neuromorphic networks ("CrossNets" [6-10]), which combine CMOS-implemented neural cells with nanoelectronic crossbars, with a continuous-state memristor at each crosspoint – see Fig. 1.

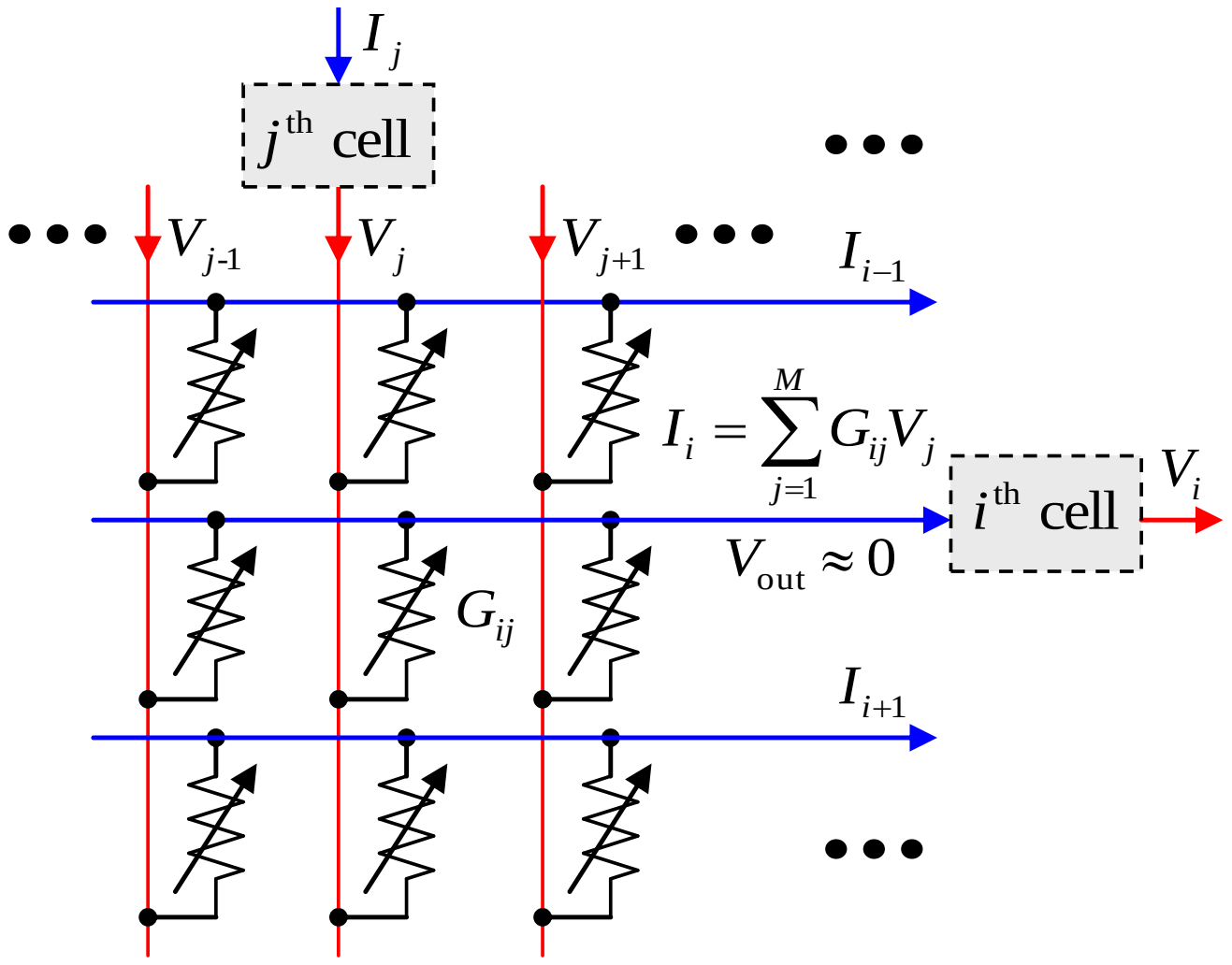

Fig. 1. The simplest memristive crossbar, which may provide adjustable, nonvolatile coupling between neural cells.

If the voltages $V_j$ applied to the crossbar input lines are not

This work was supported by a DoD MURI program via the AFOSR Award #FA9550-12-1-0038, and by the ARO under Contract #W91NF-16-1-0302. A part of numerical calculations was performed using supercomputer facilities of the DoD's HCPMP program.

* D. Gavrilov and K. K. Likharev are with Stony Brook University, Stony Brook, NY 11794, USA (e-mails: dmitri.gavrilov@stonybrook.edu; konstantin.likharev@stonybrook.edu).

** D. Strukov is with University of California at Santa Barbara, Santa Barbara, CA 93106 (e-mail: strukov@ece.ucsb.edu)

too large (for typical metal-oxide memristors, below ~1 V), they do not alter the pre-set memristor states, and the crossbar, with the virtual-ground condition $V_{out} \approx 0$ enforced on its output lines, performs a multiplication of the vector of these voltages by the matrix of memristors' conductances $G_{ij}$:

$$I_i = \sum_{j=1}^{M} w_{ij} V_j \,, \qquad (1)$$

where $I_i$ is the output current, $M$ is the cell connectivity, and $w_{ij}$ are the synaptic weights, in this simplest case proportional to $G_{ij}$. Hence the memristive crossbar performs, on the physical level, the most common operation at the inference stage of neuromorphic network operation, which is the main bottleneck at their digital implementation. As a result, the intercell communication delays in nanoelectronic CrossNets may be reduced to just few nanoseconds, and their energy efficiency may approach that of the human cerebral cortex.

The global connectivity of a limited number $N$ of neuron cells, with $M = N - 1$, may be implemented by placing the cells peripherally, around a single $(N - 1)\times(N - 1)$ crossbar. However, for most real-world applications, such global connectivity is redundant, and an area-distributed interface between a memristive crossbar and an array of CMOS-implemented neurons may be used to provide the desired restricted connectivity graph. For example, the very natural "InBar" interface topology [7] may ensure the connectivity of each neuron with all other neurons in its vicinity with a shape approaching that of a square $m\times m$, so that $M = m^2 - 1 < N$ – see Fig. 2. (For practically interesting cases, $1 << M << N$.) Such connectivity domain's shape is very convenient for the discussion of the CrossNet ASTM (though not necessary for its physical implementation), and will be used in this paper.

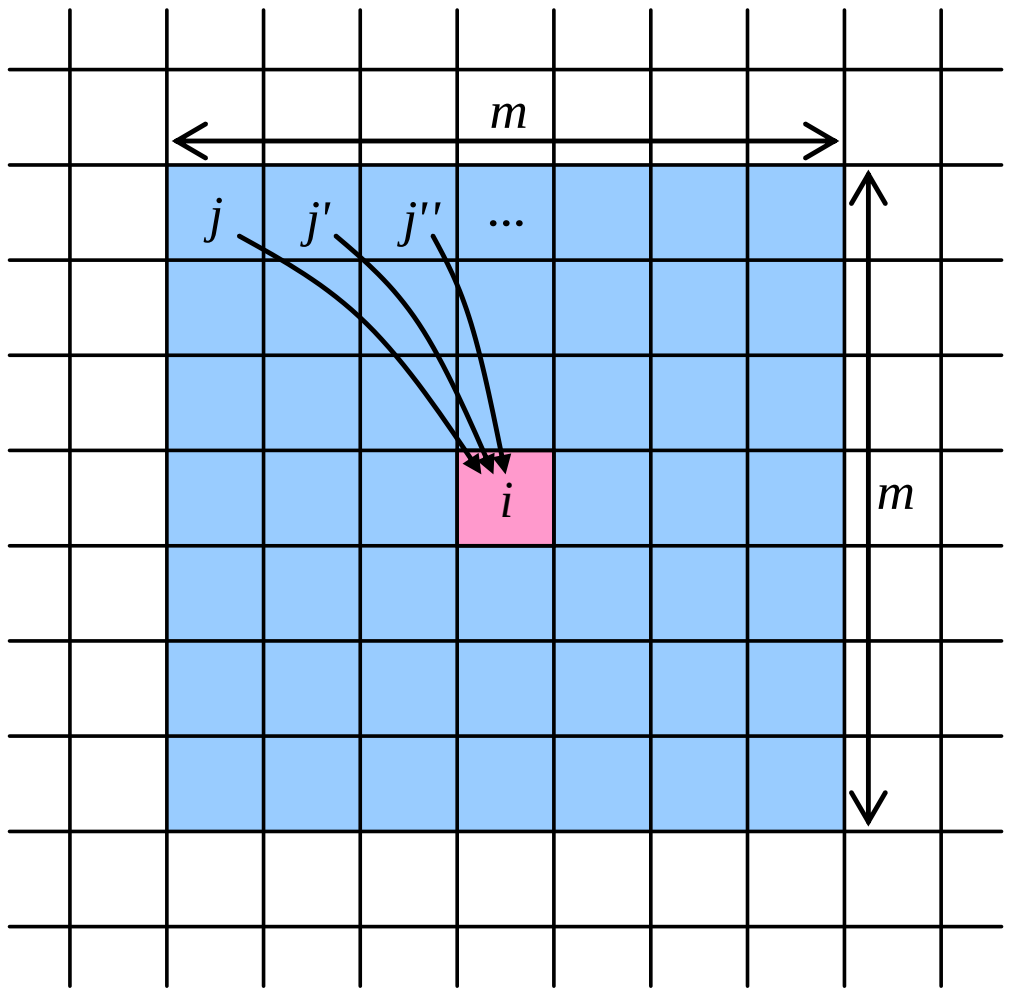

Fig. 2. The connectivity domain of a neuron cell number *i*, which may be provided by a more dense memristive crossbar with the InBar topology [7]. Note that each cell has a similar domain, so that the connection between each pair of cells is two-sided - though typically asymmetric.

Fig. 3 shows the basic idea of operation of the memory. Just as in Fig. 2, the neural cells are mapped on a rectangular grid, each of them corresponding to one B/W pixel of all movie frames. At the movie recording stage, for each pair of sequential frames, the synaptic weight connecting two pixels, within their connectivity domain, is strengthened if the two pixels that have the same value (1 or 0) in both frames, and is weakened in the opposite case. For example, in the case of Fig. 3, where the pixels of a certain sign are placed on grey background, the weights $w_{ij}$ and $w_{i'j'}$ (symbolized by solid lines) are strengthened, while the weights $w_{i'j}$ and $w_{ij'}$ (dashed lines) are weakened.

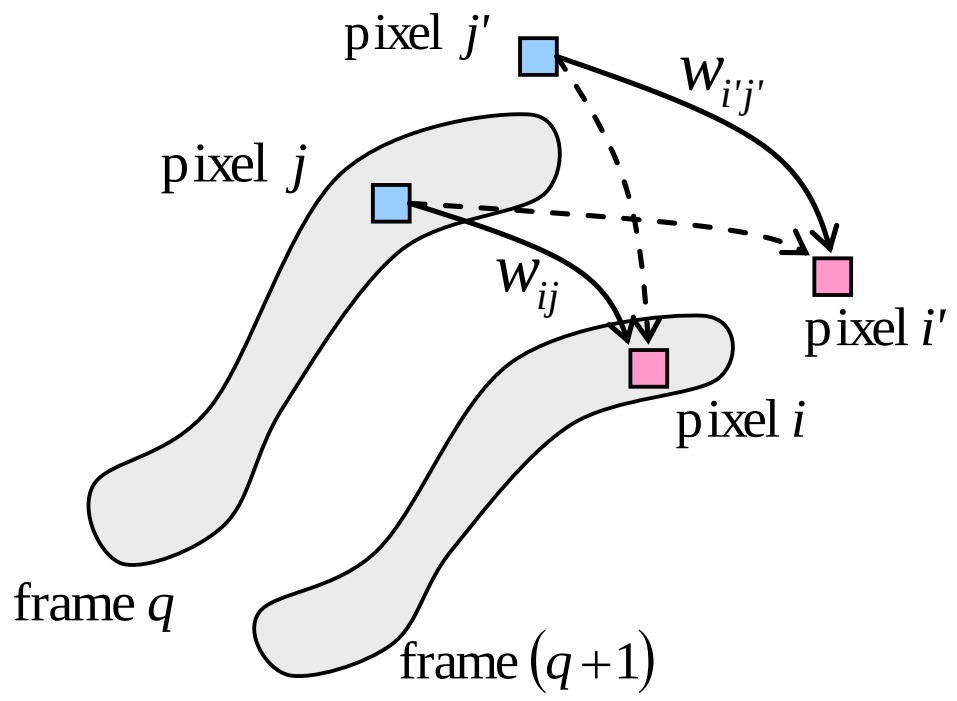

Fig. 3. The basic idea of operation of the neuromorphic ASTM.

If the recording procedure has been efficient, then at the readout (also called the "retrieval") stage, the activation of pixels by a frame leads to a correct sequential activation of the following frames of the movie, even if the input frame is incomplete, or partly corrupted by noise. However, different frame pairs typically impose contradictory requirements to the same synaptic weight $w_{ij}$, so that for each recording method, the correct retrieval is only possible if the total number $Q$ of the frames does not exceed a certain number $Q_{max}$, called the memory capacity.

The general idea of such operation of the ASTM is not quite new. Its software aspects were repeatedly discussed starting from the 1960s – see, e.g., Ref. [11]. A review of the initial work, mostly for the firing-rate networks, may be found in Sec. 3.5 of Ref. [12]. This idea was revitalized [13] at the advent of spiking network research, and in this context, discussed in quite a few publications – see, e.g., the reviews [14, 15], and the later papers [16-20]. However, to the best of our knowledge, the key issue of the ASTM capacity was addressed only in the Ph.D. thesis by S. Wills [21], for a very specific readout timing model, very inconvenient for hardware implementation. (The capacity calculated in this work is also substantially lower than for the synchronous readout discussed below.)

The objective of this work was a detailed study of the recording and readout methods, which would enable the highest capacity of the CrossNet ASTM. In Sec. II we discuss the readout options, and in particular the most critical issue of its timing. Sec. III is devoted to the description of four most plausible ways of data recording, and the calculation of the corresponding capacity-vs-fidelity tradeoffs. The results of more detailed studies of the two best recording methods, in particular of their immunity to the input frame corruption and device (memristor) variability, is discussed in Sec. IV. In the next Sec. V, the results are compared with those of the memristive T-CAM suggested in Ref. [5]. Finally, in the Conclusion (Sec. VI) we summarize our results, and discuss prospects of experimental implementation of ultrafast CrossNet ASTM.

## II. Readout Options

As follows from the above qualitative description of the memory, it is quite suitable for the asynchronous spiking mode of operation - see, e.g., Ref. [22]. In this mode, the input of all initial frame's "active" pixels (say, equal to 1) triggers simultaneous spikes $V_j(t)$ at the outputs of the corresponding neural cells. As a result of their action on the memristive crossbar, all other cells of the system receive input pulses $I_i(t)$, determined by Eq. (1). In some of the cells (ideally, all and only those corresponding to the active pixels of the next frame), the input pulses push the action potential beyond the spiking threshold, causing them to fire. This new series spikes triggers spiking in the next cell set, corresponding to the active pixels of next frame, etc.

However, in the absence of global synchronization, the cells corresponding to active pixels of a frame (besides the initial one) do not necessarily fire simultaneously, because of the previously accumulated individual action potentials. Our extensive numerical experiments, using the simple LIF model of the cells [22], have shown the following very interesting behavior. If the number $Q$ of the recorded frames is much smaller than the memory capacity $Q_{\max}$ (for a particular recording method), the frames are reproduced almost perfectly, with a relatively small time differences ("jitter") between the spikes representing each frame. As $Q$ is increased, the jitter increases substantially, but (for us, rather counter-intuitively) its intensity does not increase with each new frame, keeping its qualitative content intact. Only when $Q$ approaches $Q_{\max}$, the reproduced movie degrades into noise.

We believe that this effect may be rather interesting for theoretical neuroscience. However, we could not help noticing that the elementary global timing (synchronization) of all spikes of each frame increases the memory capacity $Q_{\max}$ rather dramatically. Such global timing may be achieved, for example, just by a periodic simultaneous lowering of the firing thresholds of all the cells, with a time period somewhat larger than the characteristic time of RC-transients in the crossbar.

Because of this, the results presented in the balance of this paper are for the globally-synchronous operation mode. In order to analyze this mode, we have used the following simple model (which blurs the difference between spiking and firing-rate operation): for each time period, corresponding to the reproduction of one frame of the movie, the voltages $V_j$ and currents $I_i$ in Eq. (1) are considered constant, with each neural cell providing a static threshold activation function $V_i^{(q+1)} = f(I_i^{(q)})$, where the upper index is the frame number ($q$ = 1, 2, …, $Q$). The activation function was taken in the simple form

$$V_i^{(q+1)} = V_0 \operatorname{sgn} I_i^{(q)}, \qquad (2)$$

which implies that $V_i$ and $I_i$ may be either positive or negative, typically leading to their zero-centered statistical distributions. Though such zero-centered operation requires differential crossbars with two memristors per synaptic weight: $w_{ij} \propto G_{ij}^{+} - G_{ij}^{-}$, it is very natural for CrossBars [7], and is also convenient for the compensation of the temperature dependence of memristor conductances - see, e.g., Ref. [23].

## III. Recording Methods: Capacity vs. Fidelity

At the first stage of our work, we have explored the tradeoff between the movie retrieval fidelity (in terms of the probability of the correct readout, in a statistical ensemble of random frames) and the network capacity $Q_{\max}$, for four most natural methods of movie recording.

### A. Hebb Rule

Conceptually, the most straightforward recording method is using the Hebb rule in its simplest form (discussed in literature as early as in 1972 [24]):

$$w_{ij} = \frac{1}{Q}\sum_{q=1}^{Q} s_i^{(q+1)} s_j^{(q)} , \qquad (3)$$

where $s_j^{(q)} = \pm 1$ are the symmetrized values of the B/W pixels of the $q^{\text{th}}$ frame. This rule evidently corresponds to the verbal description of the weight setup discussed in the Introduction, and may be implemented for *in-situ* recording using the spike-timing-dependent plasticity (STDP) – see, e.g., Ref. [25].

The capacity-to-fidelity tradeoff for this method may be readily evaluated analytically, assuming that all the binary pixels in the whole movie are random and uncorrelated. Indeed, let us assume that in a frame number $q$, all $M$ cells within the connectivity domain of an $i^{\text{th}}$ cell have correct values: $V_j^{(q)} = V_0 s_j^{(q)}$. Then plugging Eq. (3) (with the summation index replacement $q \to q'$) into Eq. (1), we may calculate the normalized product of the signal $I_j$ arriving at the $j^{\text{th}}$ cell, by its correct value, $s_i^{(q+1)}$, in the next frame:

$$\frac{Q}{V_0} I_i^{(q)} s_i^{(q+1)} = \sum_{j=1}^{M}\sum_{q'=1}^{Q} s_i^{(q+1)} s_i^{(q'+1)} s_j^{(q')} s_j^{(q)} . \qquad (4)$$

Due to the independence of different pixels, the sum of $MQ$ terms in the right-hand part of Eq. (4) has only $M$ terms (all with $q = q'$) always equal to +1, while all other terms have an equal probability to equal either +1 or -1. At $M$, $Q$ >> 1, the sum of these $M(Q-1)$ random terms has a Gaussian probability distribution with a zero statistical average, and the variance equal to $M(Q-1) \approx MQ$. As a result, the probability of the negative sign of the whole sum (4), i.e. of an error of the $i^{\text{th}}$ pixel in the $(q+1)^{\text{st}}$ frame, is

$$p \approx \frac{1}{\sqrt{2\pi MQ}} \int_{M}^{\infty} \exp\left(-\frac{x^2}{2MQ}\right) dx \equiv \frac{1}{2} \operatorname{erfc} \sqrt{\frac{M}{2Q}} , \qquad (5)$$

where erfc($x$) ≡ 1 - erf($x$) is the complementary error function.

Note that this result is similar to that for the Hopfield networks with the similarly sharp activation function (see, e.g., Sec. 2.2 in Ref. [12]), because for this calculation, the addition of 1 to the upper indices in the right-hand part of Eq. (4) is not important. (In the standard analysis of the Hopfield networks with the global coupling, $N$ - 1 plays the role of the connectivity $M$ in the restricted CrossNet.)

As a sanity check, we have verified Eq. (5) by numerical simulation for several values of $M$ and $N$. This, and all other numerical simulations described in this paper, have been performed on a square lattice of $N\times N$ neural cells. In order to mitigate the effects of large but finite size $N$, the usual cyclic boundary conditions on both pairs of opposite sides of the square have been used (equivalent to wrapping the network on a thorus). The calculation results agree with Eq. (5) within the (very small) statistical error of the numerical simulation.

In the most important limit of small error probability, Eq. (5) is reduced to

$$p \approx \sqrt{\frac{Q}{2\pi M}} \exp\left\{-\frac{M}{2Q}\right\} << 1, \quad \text{for } 1 << Q << M \cdot \tag{6}$$

At larger $p$, we need to take into account the induced errors, i.e. the effect of an error in a $q^{\text{th}}$ frame on the error probability in the $(q + 1)^{\text{st}}$ frame. At $Q$, $M >> 1$, such a calculation may be performed analytically using the mean-field approach, similar to that used for the calculation of the Hopfield network's capacity – see, e.g., Sec. 2.5 in Ref. [12]. However, since the main focus of this work was on other, better recording methods, we have opted for the simple numerical simulation of the movie retrieval. These numerical experiments have shown that at the retrieval process, the fraction of incorrect pixels per frame rapidly approaches some stationary, equilibrium value $p$; these values are plotted by points in Fig. 4 for several $N$ and $M$. As the results show, the normalized memory capacity $Q_{\max}/M$ is virtually independent of these parameters – just as in Eqs. (5) and (6).

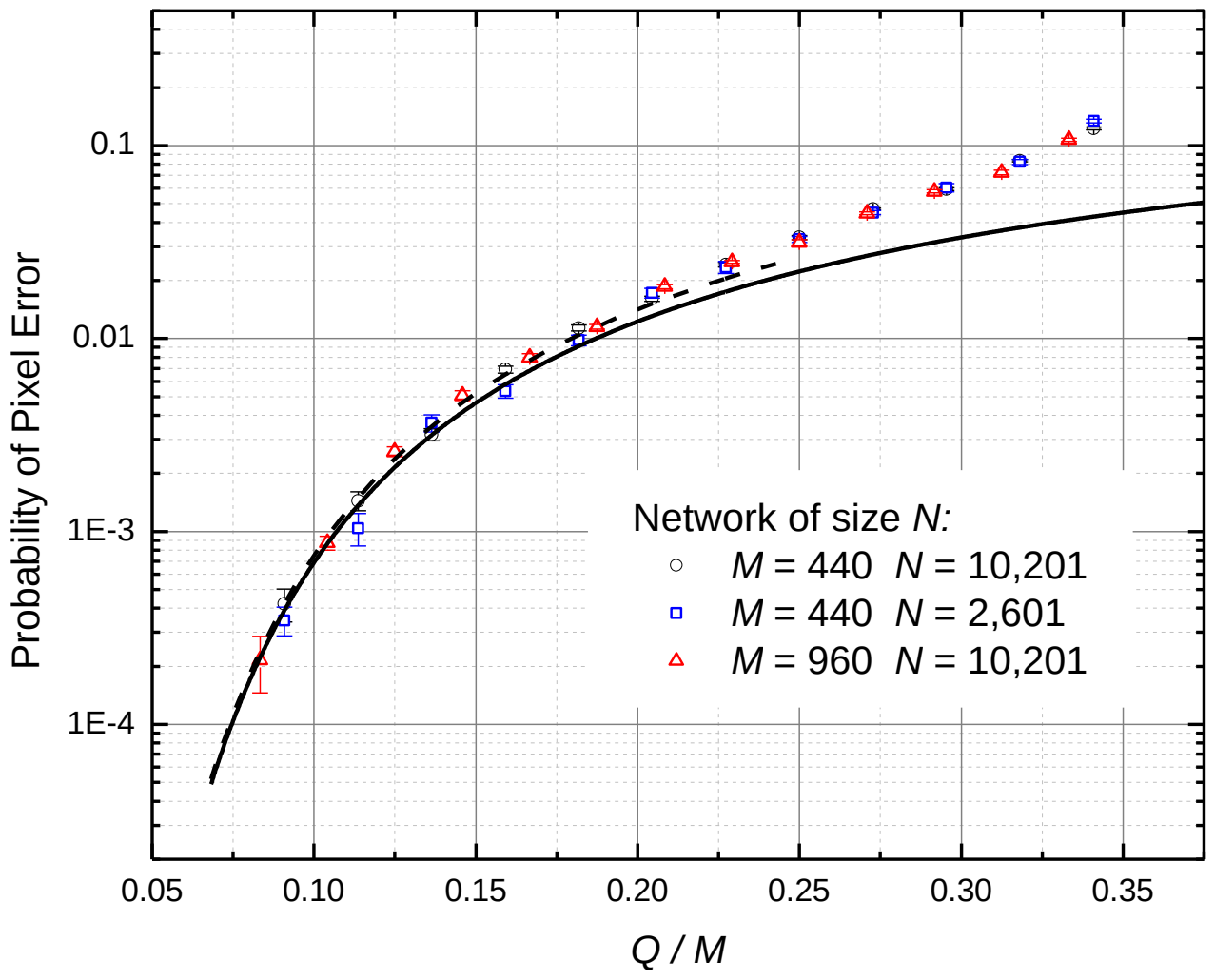

Fig. 4. The probability of pixel retrieval error in the ASTM using the Hebbian recording (3). Lower curve: Eq. (5); dashed curve: Eq. (6); upper points: numerical simulation results, which automatically take into account the induced errors.

For the practically interesting fidelity range ($p \leq 1\%$), the corrections due to induced errors are not important, and the numerical results, with a good accuracy, are described by Eqs. (5)-(6). In particular, for the 99% fidelity ($p$ = 0.01), $Q_{\max} \approx 0.18M$. Such low capacity is not too surprising, given the well-known result $Q_{\max} \approx 0.14M$ for the Hopfield networks with the similarly restricted connectivity [7], and the similar activation function (2).

## B. *Quadratic Programming*

One more natural way to calculate weights $w_{ij}$ from the given set of binaries $s_j(q)$ is to require that all $I_i^{(q)}$ have the correct sign, i.e. that of the next-frame's pixel $s_i^{(q+1)}$ :

$$I_i^{(q)} s_i^{(q+1)} \propto \sum_{j=1}^{M} s_i^{(q+1)} s_j^{(q)} w_{ij} > 0, \quad \text{for } i = 1, 2, \ldots N \,, \tag{7}$$

This system of $Q$ inequalities for each $i$ is typically insufficient to uniquely determine $M$ weights $w_{ij}$, and thus must be complemented by some reasonable additional conditions. We have first tried several available algorithms of the linear programming [26]. However, they typically lead to a growth of the width of the synaptic weight distribution, especially strong at $Q \to Q_{\max}$. Such a broad distribution is rather inconvenient for the hardware implementation, in which the range of possible memristor conductances $G$ is always limited – see, e.g., Ref. [9].

Therefore, we have moved to the quadratic programming [27], at which Eq. (7) is complemented with the requirement of the smallest norm of the vector of synaptic weights $w_{ij}$. The calculations have been performed using the MATLAB's function *quadprog()*. This procedure takes much more significant computing resources than the previous (Hebb-rule) recording method.

The simulations have shown that with the growth of the number $Q$ of the recorded frames, the retrieval proceeds differently than at the Hebb-rule recording. Namely, the number of wrong pixels in each retrieved frame is typically very small, but when a few errors appear, they almost immediately lead to a compete corruption of the remaining frames of the movie. As a result, the system's fidelity violation is better characterized by the probability $p$ of the movie corruption, measured on a large statistical ensemble of different movies (again, with completely random and independent pixels).

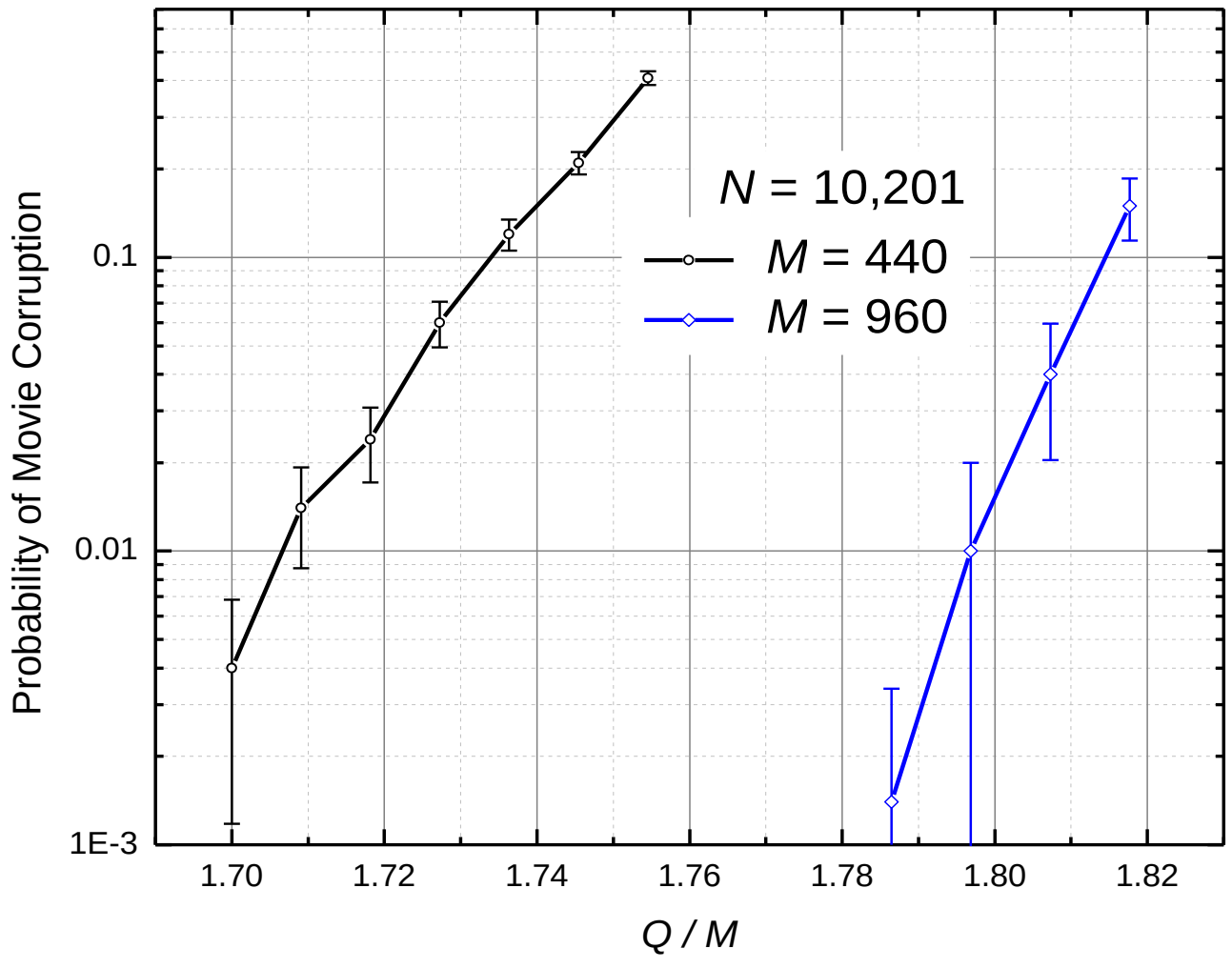

Fig. 5. The numerically simulated probability of retrieved movie corruption in the ASTM using the quadratic programming. (The curves are only guides for the eye.) The error bars represent standard deviation of the mean based on 500 simulations for $M$ = 440 and 100 simulations for $M$ = 960.

Fig. 5 shows the $p$ so defined as a function of the same ratio $Q/M$ as in Fig. 4. The results show that for a reasonable fidelity (say, $p$ = 1%), the network capacity $Q_{max}$ is close to ~(1.75 ± 0.05)$M$, i.e. is almost an order of magnitude higher than for the Hebb-rule recording.

Note that for the case of global connectivity ($M = N - 1$), this number is close to the theoretical capacity maximum $Q_{max}=2(N - 1)$ of the usual (spatial) associative memory, based on a recurrent neuromorphic network [28].

### C. Analog Gradient Descent

The next natural recoding method is an iterative algorithm similar to the well-known delta-rule of the feedforward perceptron training, describing the gradient descent of the quadratic error function – see, e.g., Sec. 5.4 of Ref. [12]:

$$\Delta w_{ij} = -\eta s_j^{(q)} \varepsilon_i^{(q+1)}. \qquad (8)$$

Here $\eta$ is a (small) training rate, and $\varepsilon$ is the error of the previous prediction of the next frame's pixel:

$$\varepsilon_i^{(q+1)} = \sum_{j=1}^{M} w_{ij} s_j^{(q)} - s_i^{(q+1)}. \qquad (9)$$

The numerical simulation has shown that after such a recording, the movie retrieval dynamics is qualitatively similar to that at the quadratic programming (see the previous subsection): an increase of the number $Q$ of the recorded frames leads to an increase of the probability $p$ of the total corruption of the retrieved movie. Fig. 6 shows a typical dependence of this probability on the ratio $Q/M$; it indicates that the memory's capacity is approximately twice lower than that for the quadratic-programming recording; for $p$ = 1%, $Q \approx 0.97M$.

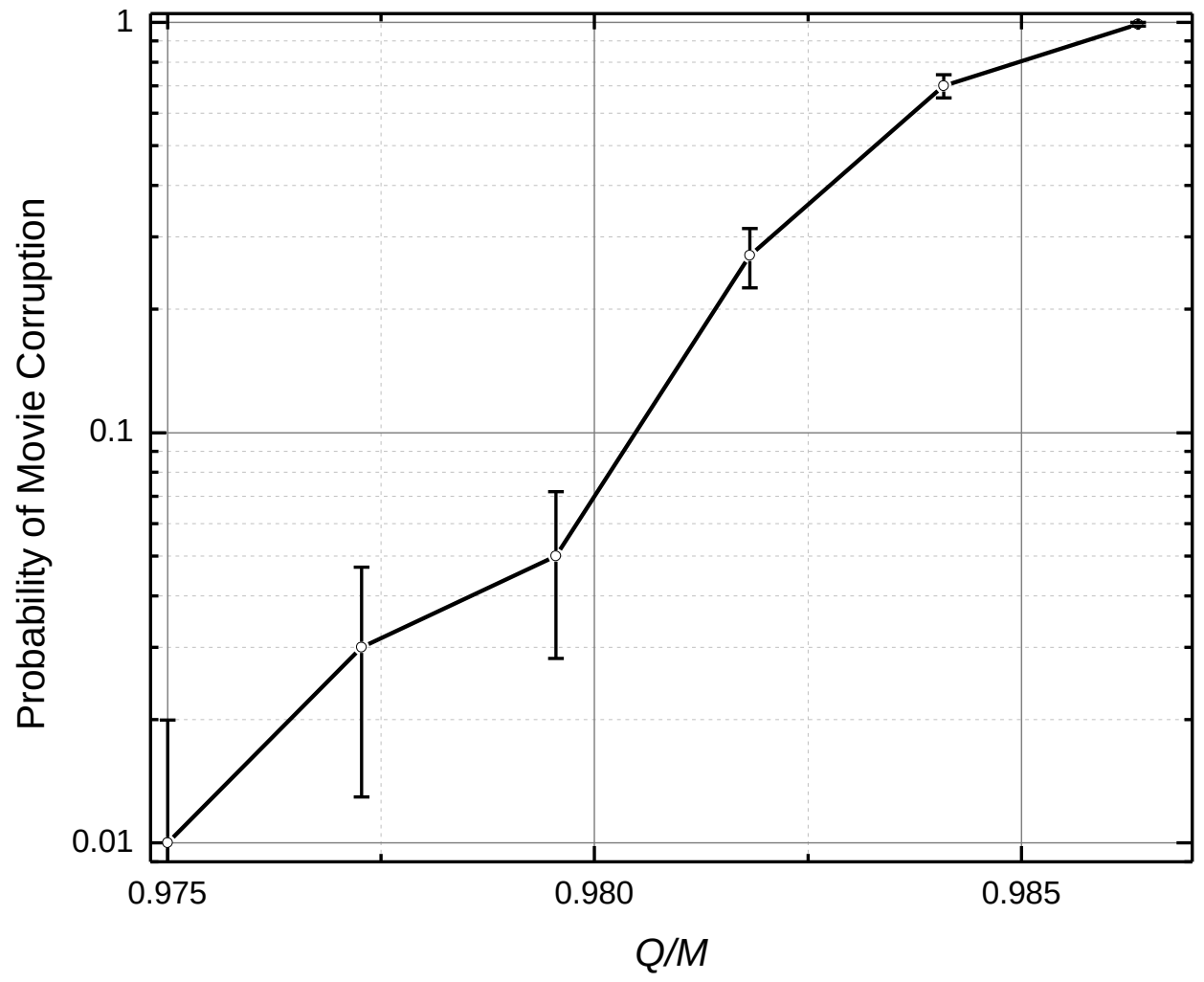

Fig. 6. The probability of the retrieved movie corruption in an ASTM using the analog gradient-descent recording. (The lines are only guides for the eye.) $N = 101\times101 = 10{,}201$, $M = 21\times21 - 1 = 440$, $\eta = 10^{-3}$. The iterations (8) were stopped either after $10^5$ epochs, or when the magnitude of all errors (9) dropped below 0.1. The error bars represent standard deviation of the mean based on 100 simulations.

In hindsight, such relatively poor results might be anticipated. Indeed, the algorithm (8)-(9) forces the network outputs to approach the *exact* integer values $s_i^{(q+1)}$ of the next pixels, while for the successful movie retrieval, it is only necessary for it to have its sign correct – see Eq. (2). As the result, the unnecessary changes of the weights interfere with the substantial ones, and hinder the iterations' efficiency.

### D. Discrete Gradient Descent

The analog gradient descent method may be improved just by rounding ("clipping") the sum in Eq. (9) to the closest of ±1. We have found, however, that even better results may be obtained by the following modification of this relation:

$$\varepsilon_i^{(q+1)} = S_i^{(q+1)} - s_i^{(q+1)}, \qquad (10)$$

where the integer $S$ depends not only on the current prediction of the output pixel, as in Eq. (9), but also on its proper value:

$$S_i^{(p+1)} = \mathrm{sgn}\left( \sum_{j=1}^{M} w_{ij} s_j^{(q)} - D s_i^{(q+1)} \right), \qquad (11)$$

where $D$ is some phenomenological parameter, to which the results are not too sensitive. (After some experimenting, we have finally selected $D$ = 1).

Fig. 7 shows the probability of the retrieved movie corruption as a function of the normalized number $Q$ of the recorded frames, for several values of parameters $N$ and $M$. The results imply that the capacity-to-fidelity tradeoff is almost as good as that available from the (much less convenient) quadratic programming; for example, at $p$ = 1%, $Q_{max} \approx (1.67 \pm 0.02)M$, depending on $M$ and $N$.

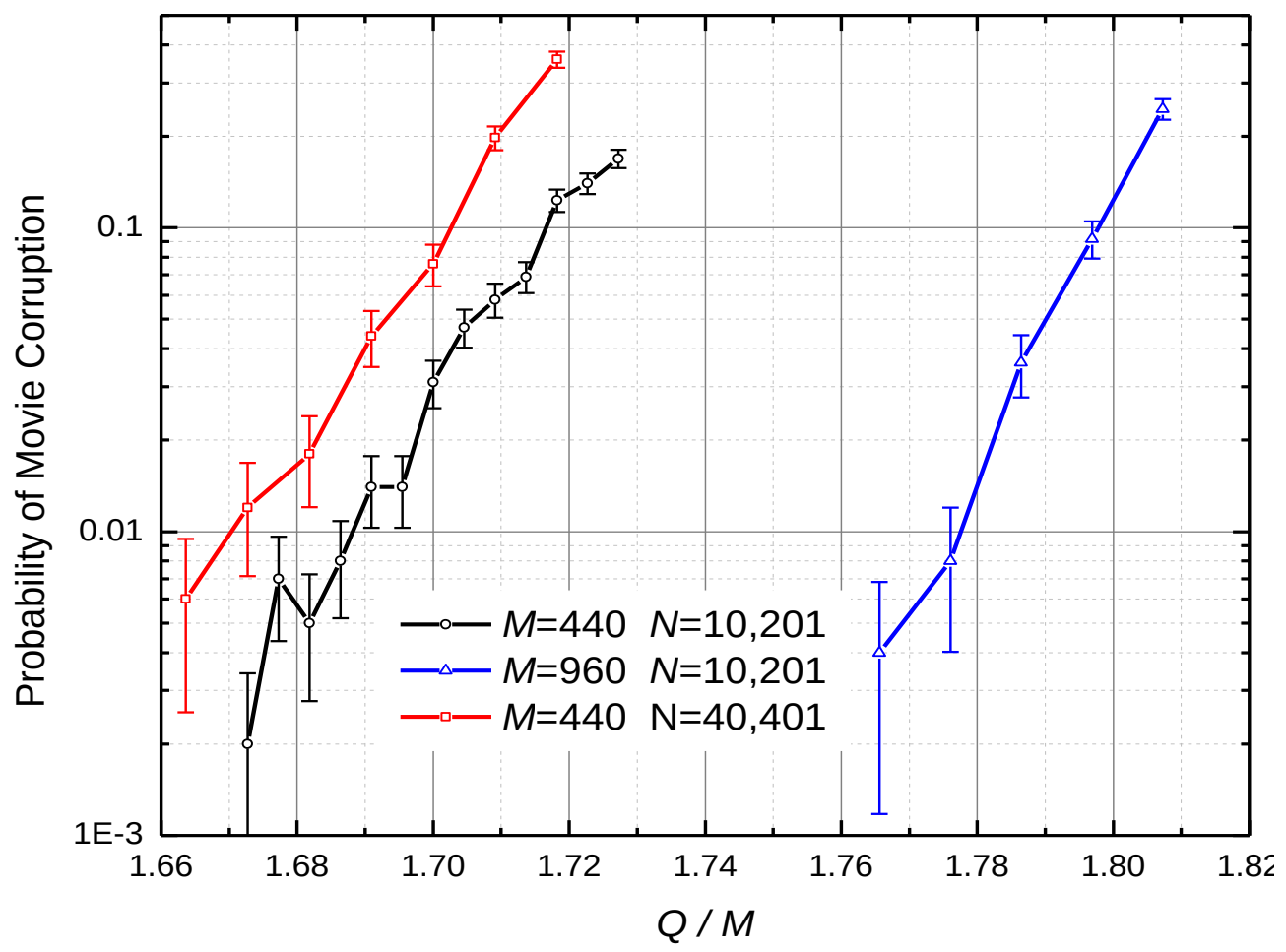

Fig. 7. The probability of retrieved movie corruption in the ASTM using the discrete gradient-descent recording described by Eqs. (8), (10), and (11). (The lines are only guides for the eye.) $\eta$ = 0.01; $D$ = 1. The iterations were stopped when the errors $\varepsilon_i^{(q+1)}$, defined by Eq. (10), reached 0 for all $i$ and $q$. The error bars represent the standard deviation of the mean, based on 1,000 simulations for $M$ = 440, and 500 simulations for $M$ = 960.

Note that all the CrossNet ASTM capacity results, shown in Figs. 4-7, are for completely random binary (B/W) pixels, i.e. for the 50% probability for each pixel to have a certain value (±1). If this probability is either lower or higher, the capacity is even larger – see, e.g., the results shown in Fig. 8.

For very sparse patterns (with either $d \ll 1$ or $1 - d \ll 1$), even higher capacity may be possible using a natural modification of the recording rules suggested for usual (spatial) associative memories – see, e.g., pp. 52-53 in Ref. [12]. At the software implementations of the memories, these rules are sometimes applied to dense patterns (with $d \sim ½$) as well, using their mapping on sparse ones. At the hardware implementation, however, such approach would require an impracticable increase of the necessary resources.

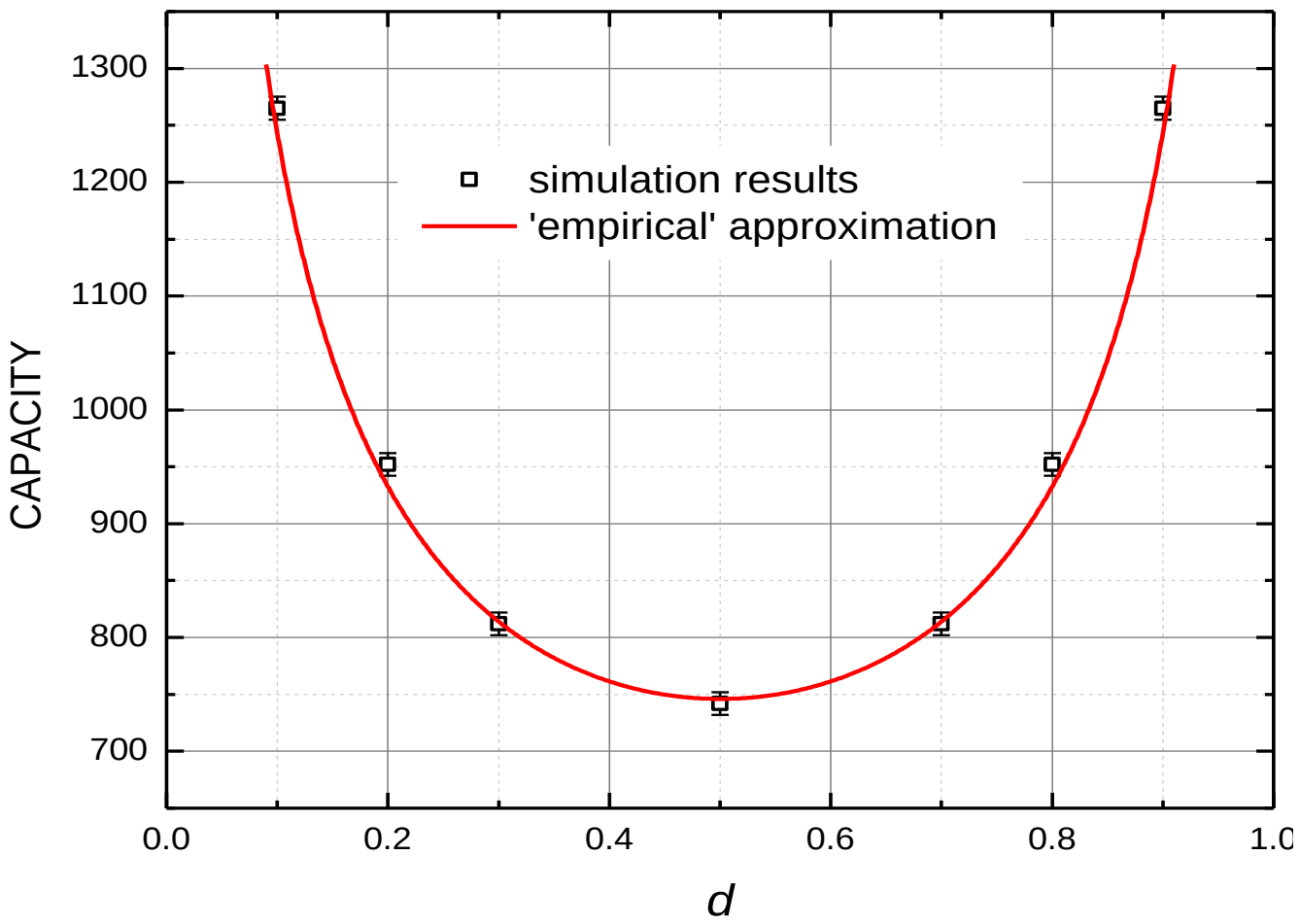

Fig. 8. The capacity (at 99% fidelity) of the ASTM using the discrete gradient-descent recording, as a function of the "duty cycle" $d$, i.e. the fraction of binary pixels having a certain value (+1) in each frame, for $M$ = 440. The smooth curve shows the empirical dependence $Q_{max} \propto 1/[d(1 - d)]^{1/2}$. The error bars represent estimated maximum deviation.

## IV. Immunity to Noise and Device Variability

To summarize the previous section, two of the methods we have studied, *B* and *D*, stand out of the competition: the former one (based on the Quadratic Programming) due to the largest memory capacity, and the latter one (based on a discrete version of the gradient descent approach) due to its local nature, enabling hardware implementation of the recording, with a minimal involvement of peripheral circuitry - at a very competitive capacity. These two methods have been chosen for a more detailed study, namely a numerical evaluation of the CrossBar ASTM's immunity to the noise contamination of the input frame, and of its tolerance to random deviations of the synaptic weights from the optimal values calculated at the recording. (Random deviations of weights were simulated by adding random deviations to the original weights before each movie retrieval attempt. The deviations were random and independent, obeying the Gaussian distributions with zero mean, and a relative r.m.s. value *r*.)

The results of these calculations are presented, respectively, in Figs. 9 and 10. The plots in Fig. 9 show, for example, that if the number $Q$ of movies recorded into an ASTM, by either of the two methods, is 25% of its maximum capacity, it may recognize the input frame with ~10% corrupted pixels, but if $Q$ is increased to 50% of $Q_{max}$, the input noise tolerance drops sharply, to only $\sim 10^{-3}$ of the pixels.

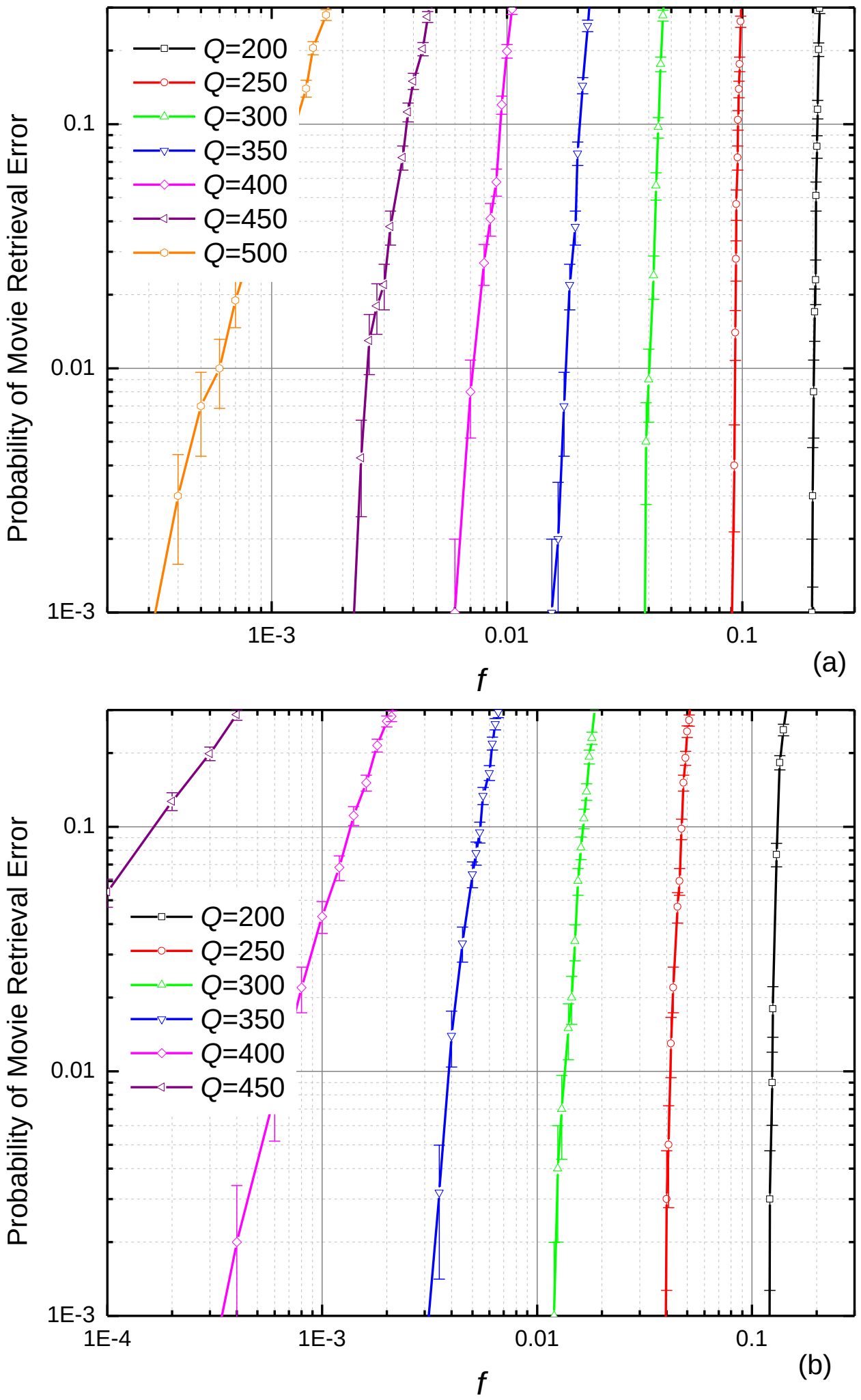

Fig. 9. The probability of movie retrieval error in the ASTMs using (a) the quadratic-programming recording, and (b) the discrete gradient-descent recording, as functions of the fraction $f$ of wrong (randomly flipped) binary pixels in the input frame, for $M$ = 440. The error bars represent standard deviation of the mean based on 1,000 simulations.

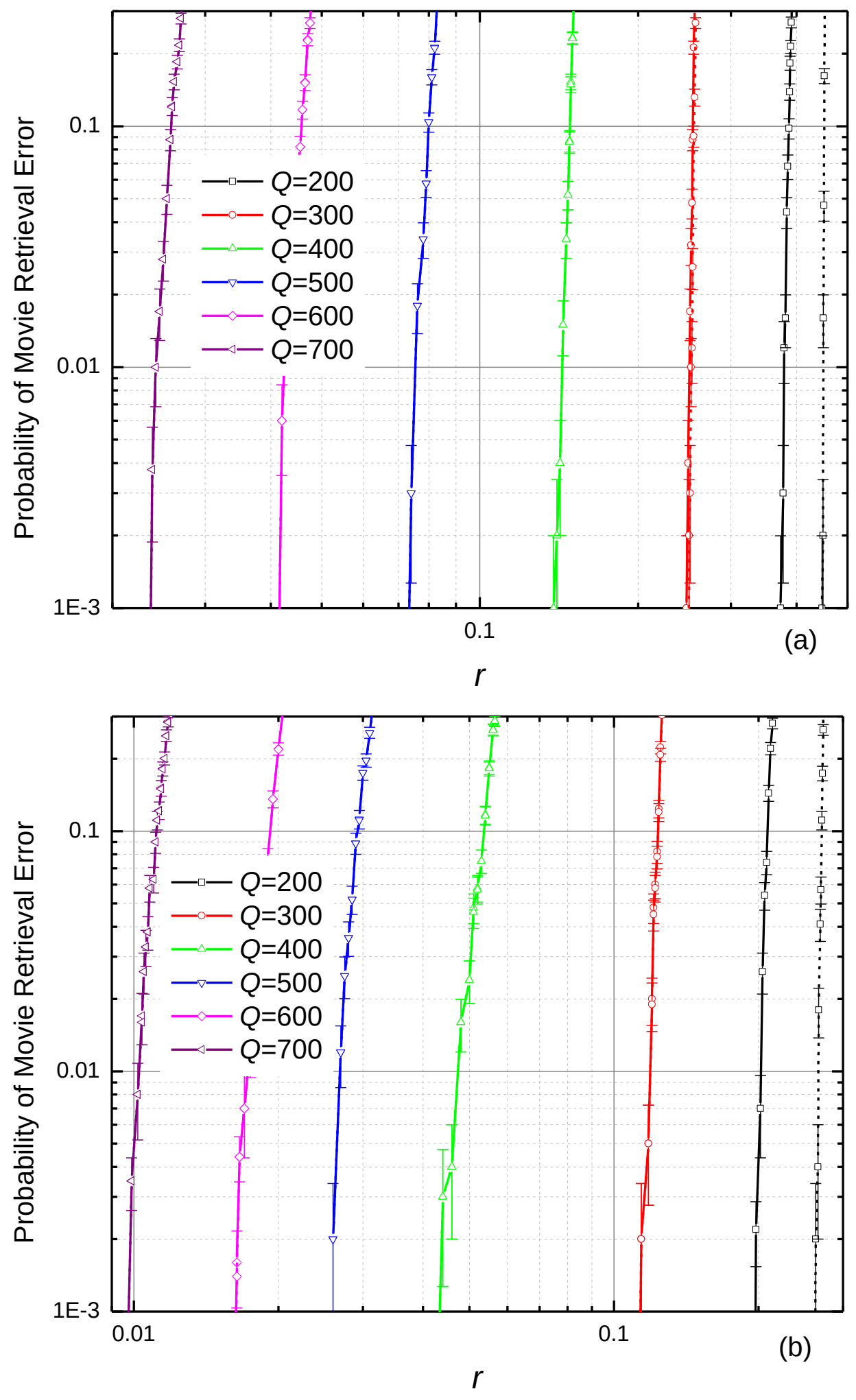

Fig. 10. The probability of movie retrieval error in the ASTMs with $M$ = 440, using (a) the quadratic-programming recording, and (b) the discrete gradient-descent recording, as functions of the normalized (relative) r.m.s. deviation $r$ of the synaptic weights from the optimal (calculated) values. The error bars represent standard deviation of the mean based on 1000 simulations.

On the contrary, as Fig. 10 shows, the effect of the memory filling on its memristor fluctuation tolerance is much more smooth. For example, if the ASTM is filled to 25% of its full capacity, its operation is not hindered by ~30% weight fluctuations, while an increase of the filling to 50% reduces the tolerated r.m.s. fluctuation only to ~15%.

It is important to note that these results characterize not an instant, but rather a gradual suppression of the input noise – or the failure of thereof. For example, Fig.11 shows the number of wrong pixels in $N$ = 10,201-pixel frames for 10 simulated movie retrievals, for a system with the cell connectivity $M$ = 440, with $Q$ = 250 frames recorded using the discrete gradient descent method. The plots show that all 500 input errors (which were independent for each retrieval attempt) eventually disappeared in 8 cases, but led to a full movie corruption in two cases. These data are a small part of a 1,000 movie set which gave the point with $f \equiv 500/10{,}201 \approx 0.049$ and the error probability ~0.2, shown in Fig. 9b.

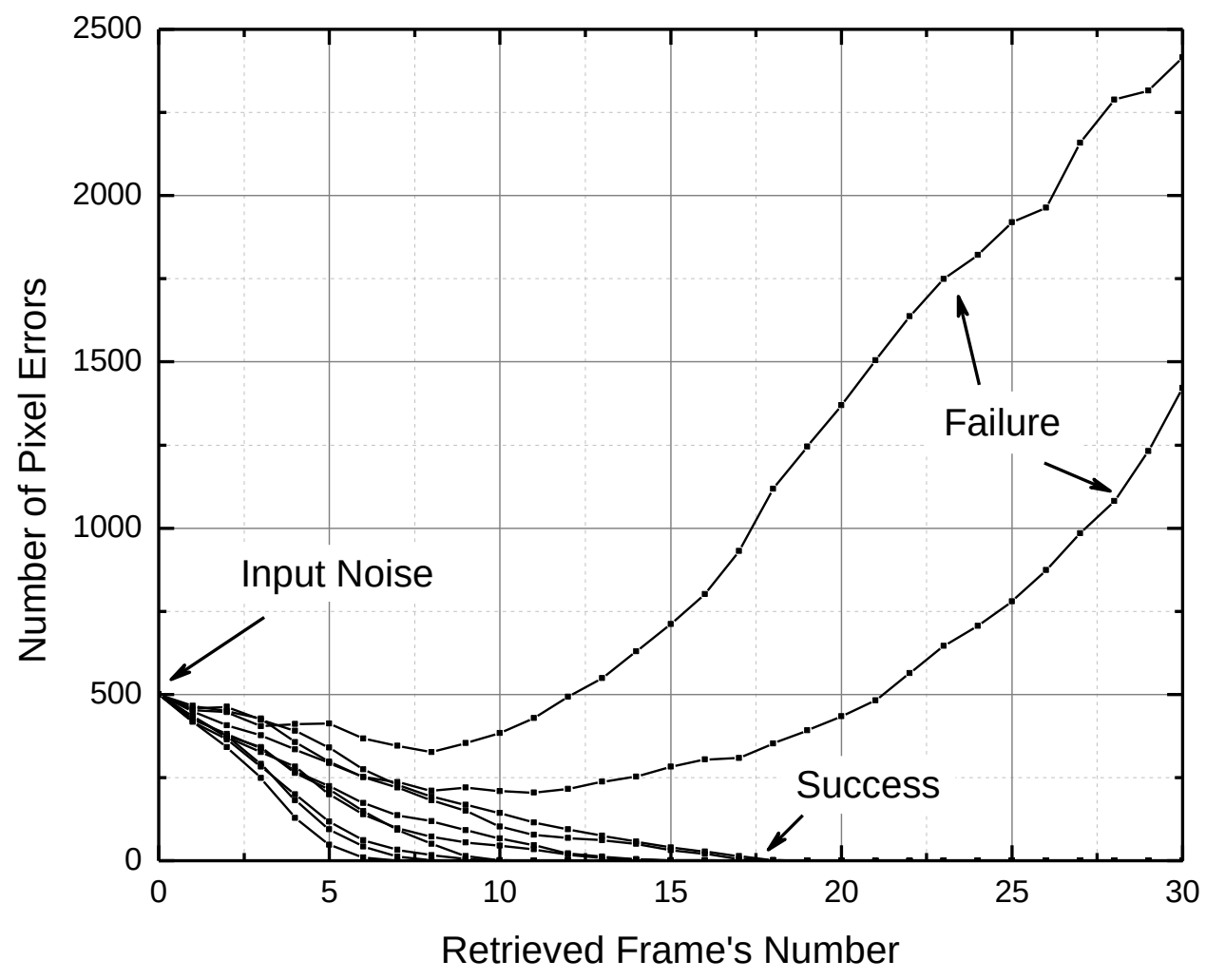

Fig. 11. The input noise suppression by a CrossNet with $N$ = 101×101 = 10,201; $M$ = 21×21 – 1 = 440 in the process of movie retrieval, as simulated for 10 independent random noise patterns. The recording of the movie with $Q$ = 250 frames was performed using the discrete gradient descent method.

## V. Comparison with T-CAM

The results shown in Figs. 9 and 10 need to be compared with those for the main competitor of the CrossNet ASTM, the T-CAM circuits already mentioned in the Introduction. Fig. 12 shows a 2×3-cell fragment of the memristor-based T-CAM [5]. It is a rectangular matrix of cells, with two binary-state memristors (plus two diodes) per cell, with each bit stored in the complimentary binary states (ON and OFF) of these two memristors, in the order encoding the bit.

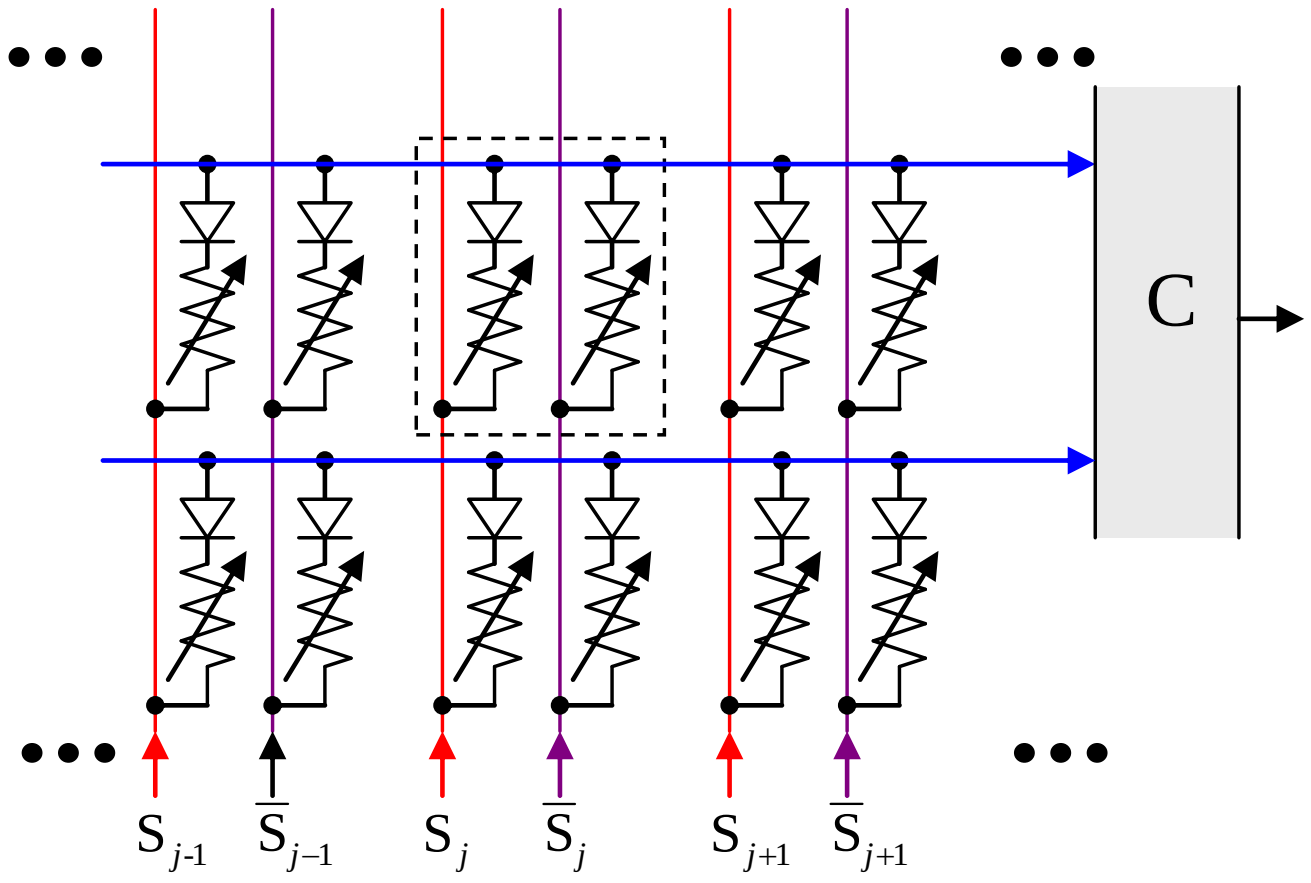

Fig. 12. ASTM implemented as a memristor-based Ternary Content-Addressable Memory (T-CAM).

In the ASTM application (again in the movie language), the $N$ binary pixels of each frame are stored in one row of the matrix, so that storage of $Q$ frames requires $Q$ rows. Before the movie retrieval, the row lines are pre-charged to the same voltage $V_0$. The retrieval is induced by feeding each pair of column lines with voltages $\{V_0, 0\}$, in the order dictated by the value of the corresponding binary pixel of the input frame. If the recorded bit of a cell corresponds to the input bit (i.e., if the input voltage $V_0 > 0$ corresponds to the ON state of the corresponding memristor, with a high conductance, while the input voltage 0, to the OFF state, with its very low conductance), the feed does not result in a noticeable current through the cell. As a result, if a recorded frame exactly matches the input one, the row line's voltage stays high. On the contrary, if some bits of a recorded frame are different from those of the input frame, the corresponding row line discharges, with the rate proportional to number of the misfit bits, i.e. to the Hamming distance between these two-bit strings. The discharge rates of all rows are compared by the comparator C, and the row with the slowest rate is assumed to carry the requested frame. After the choice of the row has been made, the whole movie may be played out without any further input. (This design may be readily generalized to more than two dimensions – see, e.g., Ref. [29].)

The fact that this circuit requires $n = 2NQ$ memristors (besides the diodes, and peripheral circuits including the comparator) may be represented by saying that the frame capacity of the T-CAM with $n$ memristors is

$$Q_{\max} = \frac{n}{2N}. \tag{12}$$

This value should be compared with the best result $Q_{\max} \approx 2M$ for the CrossNet ASMT discussed in this paper. Since in that memory, with the differential encoding of the synaptic weights, the total number of memristors is $n = 2MN$, that result may be rewritten as

$$Q_{\max} \approx \frac{n}{N}, \tag{13}$$

i.e. the capacity (12) of the T-CAM with the same number of memristors is approximately twice lower, even with two memristor per weight.

If the frame of $N$ binary pixels, submitted to T-CAD for the recognition, has some number (say, $fN$) of corrupted pixels, there is a chance that its Hamming distance from a wrong recorded frame will be lower that that from the correct frame, so that the memory will recall that wrong frame. Since the Hamming distance between two random strings, of $N >> 1$ bits each, obeys the Gaussian distribution with the mean $N/2$ and the variance $N/4$, the probability of such an error is

$$p = \frac{1}{\sqrt{2\pi(N/4)}} \int_0^{fN} \exp\left[-\frac{(k-N/2)^2}{2(N/4)}\right] dk \equiv \frac{1}{2}\left\{\text{erf}\left[(2f-1)\sqrt{\frac{N}{2}}\right] - \text{erf}\left(-\sqrt{\frac{N}{2}}\right)\right\}. \tag{14}$$

According to this formula, at $N >> 1$ the error is extremely small until the fraction $f$ of the pixels in the input frame approaches 50% very closely - by the distance of the order of $1/(2N)^{1/2} << 1$. Hence, the noise immunity of the T-CAM is higher than that of the CrossNet ASMT – cf. Fig. 9.

The memristor fluctuation tolerance of the T-CAM is also higher than that in the CrossNet ASTM. In order to calculate it, we should take into account that the Ohmic conductance $G$ of real-life memristors is nonvanishing even in the OFF state. Hence the voltage decay rate in the line corresponding to the perfect fit to the input frame (Fig. 12) is $NV_0G_{\text{OFF}} > 0$. On the other hand, the average rate of a misfit line discharge is $NV_0G_{\text{ON}}/2$, with an r.m.s. fluctuation scaling as $\sqrt{N} << N$. Hence an error due to the worst-case (simultaneous) fluctuations of memristor conductances appears only at

$$(G_{\text{OFF}})_{\max} > \frac{1}{2}(G_{\text{ON}})_{\max} \tag{15}$$

- the situation highly unlikely even at the current, immature state of the memristor fabrication technology.

## VI. Conclusion

Our calculations have shown that the CrossNet-based associative spatial-temporal memories, with appropriate methods of information recording, may be more hardware-saving than the alternative, T-CAM circuits of the same capacity, though their input noise immunity and memristor variability tolerance are lower. It is important to note, that ASTM's capacity increases naturally, without any modifications to the network, for more realistic cases of correlated frames (see Fig.8). On the other hand, T-CAM implementations would have to rely on coding and/or compression algorithms, which might have substantial implementation overhead and inferior information capacity. One more challenge for the experimental implementation of the CrossNet ASCM is the still immature technology of memristive crossbar hybridization with underlying CMOS circuits [30]; note, however, that in fully-connected CrossNets (with $M = N - 1$) the CMOS circuits may be placed peripherally, making the integration easier.

Also, the field of possible applications our results is much broader than the memristor technology. For example, they are fully applicable to CrossNets using floating-gate memory cells, with analog data recording, as synapses – see, e.g., Ref. [31]. At the industrial-grade implementation of such cells, they may be quite comparable with memristors in size, and provide almost similar speed and energy efficiency [32]. The recent fast progress of experimental work in this direction [32-33] gives every hope that the CrossNet ASTMs based on such technology may become valuable components of future ultrafast cognitive hardware systems.

## Acknowledgments

The authors are grateful to E. A. Feinberg for useful discussions.

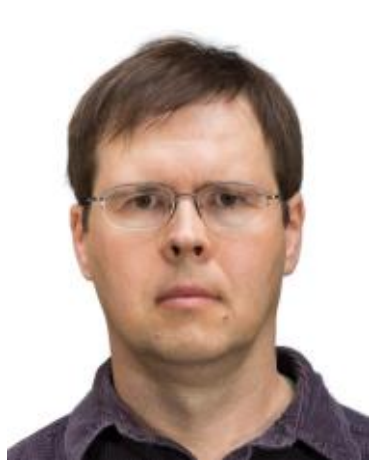

**Dmitri Gavrilov** (M'94) received the M.S. degree in electrical engineering from Saint-Petersburg State Academy of Aerospace Instrumentation, Saint-Petersburg, Russia, in 1994, and the Ph.D. degree in electrical engineering from Stony Brook University, Stony Brook, NY, USA, in 2002. Since 2002 he is working as Postdoctoral Associate, and since 2005 as a Research Scientist, at the Department of Electrical and Computer Engineering at Stony Brook University. His research interests include the areas of signal processing, pattern recognition and machine learning with applications to sensor technology.

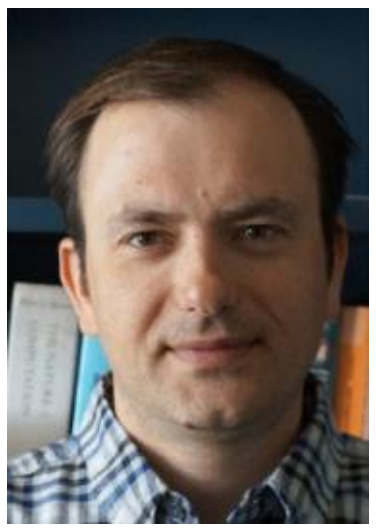

**Dmitri B. Strukov** (M'02 – SM'16) received the M.S. degree in applied physics and mathematics from the Moscow Institute of Physics and Technology, Moscow region, Russia, in 1999, and the Ph.D. degree in electrical engineering from Stony Brook University, Stony Brook, NY, USA, in 2006. He is currently a Professor of Electrical and Computer Engineering at University of California at Santa Barbara. Prior to joining UCSB Dmitri worked as a postdoctoral associate, first at Stony Brook University (Aug. 2006 – Dec. 2007), and then at Hewlett Packard Laboratories (Jan. 2007 – Jun. 2009) on various aspects of nanoelectronic systems. His research broadly concerns different aspects of computation, in particular addressing questions on how to efficiently perform computation on various levels of abstraction. His current research focus is on hardware implementations of artificial neural networks with emerging memory devices.

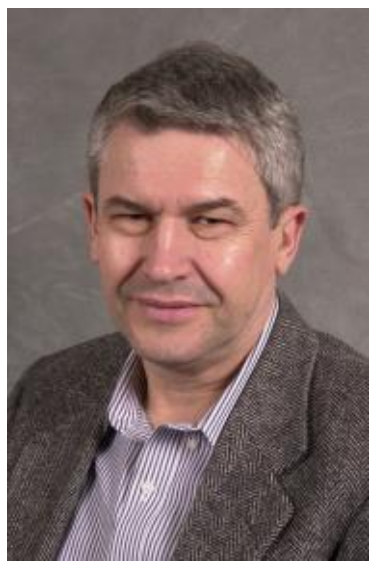

**Konstantin K. Likharev** (M'91-SM'06-F'08) received the Ph.D. in physics from Moscow State University (Moscow, USSR) in 1969, and the habilitation degree of Doctor of Sciences from the same university and the USSR Highest Attestation Committee in 1979.

From 1969 to 1988 he was a Staff Scientist of Moscow State University, and from 1989 to 1991 the Head of the Laboratory for Cryoelectronics of that university. In 1991 he assumed a Professorship at Stony Brook University – a State University of New York (Distinguished Professor since 2002). During his research career, he worked in the fields of nonlinear classical and dissipative quantum dynamics, and solid-state physics and electronics, notably including superconductor electronics and nanoelectronics. His current research interests are focused on the architecture and nanoelectronic implementation of high-performance neuromorphic networks.

Prof. Likharev is an APS Fellow; he is an author of more than 350 original publications, 75+ review papers and book chapters, 2 monographs, and several patents.